\documentclass[journal]{templates/IEEEtran}

\usepackage[utf8x]{inputenc}
\usepackage[T1]{fontenc}
\usepackage[english]{babel}
\usepackage{amsmath}
\usepackage{amsfonts}
\usepackage{amssymb}
\usepackage{amsthm}
\usepackage{bm}
\usepackage{xspace}
\usepackage{caption}
\usepackage{subcaption}
\usepackage[boxruled]{algorithm2e}
\usepackage{mathrsfs}
\usepackage{multirow}
\usepackage{eurosym}
\usepackage{dsfont}
\usepackage[normalem]{ulem} 
\usepackage{balance}
\usepackage{soul}
\usepackage{pdfpages}
\usepackage{graphicx}
\usepackage{todonotes}
\usepackage{url}
\usepackage{siunitx}
\usepackage{pgfplots}
\usepackage[hidelinks]{hyperref}
\usepackage{stfloats}
\usepackage{array}
\usepackage{adjustbox}
\usepackage{float}
\usepackage{enumitem}
\setitemize{leftmargin=*}
\usepackage{balance}
\usepackage{cite}
\newcolumntype{L}[1]{>{\raggedright\let\newline\\\arraybackslash\hspace{0pt}}m{#1}}
\newcolumntype{C}[1]{>{\centering\let\newline\\\arraybackslash\hspace{0pt}}m{#1}}
\newcolumntype{R}[1]{>{\raggedleft\let\newline\\\arraybackslash\hspace{0pt}}m{#1}}

\newcommand{\T}{\mathsf{T}}
\newcommand{\norm}[1]{\left\lVert#1\right\rVert}
\newcommand{\R}{\mathbb{R}}

\newcommand{\SO}{SO(3)}
\newcommand{\SE}{SE(3)}
\newcommand{\mat}[1]{\begin{bmatrix}#1\end{bmatrix}}
\newcommand{\bmd}[1]{\bm{\dot{#1}}}

\title{Collaborative Bimanual Manipulation Using Optimal Motion Adaptation and Interaction Control}

\author{Ruoshi Wen, Quentin Rouxel, Michael Mistry, Zhibin Li, Carlo Tiseo
\thanks{Ruoshi Wen, Michael Mistry, and Carlo Tiseo are with the Institute for Perception, Action, and Behaviour, School of Informatics, University of Edinburgh, UK. Zhibin Li is with the Department of Computer Science, University College London, UK. Carlo Tiseo is with the School of Engineering and Informatics, University of Sussex, UK. For the purpose of open access, the author has applied a Creative Commons Attribution (CC BY) license to any Author Accepted Manuscript version arising.}}

\begin{document}

\maketitle

\begin{abstract}
This work developed collaborative bimanual manipulation for reliable and safe human-robot collaboration, which allows remote and local human operators to work interactively for bimanual tasks. We proposed an optimal motion adaptation to retarget arbitrary commands from multiple human operators into feasible control references. The collaborative manipulation framework has three main modules: (1) contact force modulation for compliant physical interactions with objects via admittance control; (2) task-space sequential equilibrium and inverse kinematics optimization, which adapts interactive commands from multiple operators to feasible motions by satisfying the task constraints and physical limits of the robots; and (3) an interaction controller adopted from the fractal impedance control, which is robust to time delay and stable to superimpose multiple control efforts for generating desired joint torques and controlling the dual-arm robots. 
Extensive experiments demonstrated the capability of the collaborative bimanual framework, including (1) dual-arm teleoperation that adapts arbitrary infeasible commands that violate joint torque limits into continuous operations within safe boundaries, compared to failures without the proposed optimization; (2) robust maneuver of a stack of objects via physical interactions in presence of model inaccuracy; (3) collaborative multi-operator part assembly, and teleoperated industrial connector insertion, which validate the guaranteed stability of reliable human-robot co-manipulation.
\end{abstract}


\section{Introduction}\label{sec:introduction}
Human-robot collaboration (HRC) technique allows humans and robots to work collaboratively in a shared workspace to complete a task, and it has played an important role in Industry 4.0 \cite{industry}, space exploration \cite{spacexplore} and medical applications \cite{Tiseo2021HapFic}. Compared to the progress made in pick-and-place, grasping and pushing tasks by single-arm manipulation, dual-arm manipulation has its unique advantages in moving tasks, such as heavy, bulky and large objects, and other complex tasks in manufacturing settings such as part assembly, connector insertion, etc. However, motion planning in complex long-horizon manipulation tasks is difficult for high degree-of-freedom (DoF) robots when they have different interaction modes which involve complex contact dynamics and friction, so autonomous bimanual manipulation tasks still remain open challenges in robotics. Efficient collaboration between human operators and robots will combine the advantages from both sides, i.e., human dexterity, flexibility and adaptability, and robots' high payload, guaranteed accuracy and 24/7 running. Therefore, the collaborative bimanual manipulation framework is worth investigating to resolve the challenges in bimanual manipulation.

\begin{figure}[t]
    \centering
	\includegraphics[trim=0cm 0cm 0cm 0cm,clip,width=\linewidth]{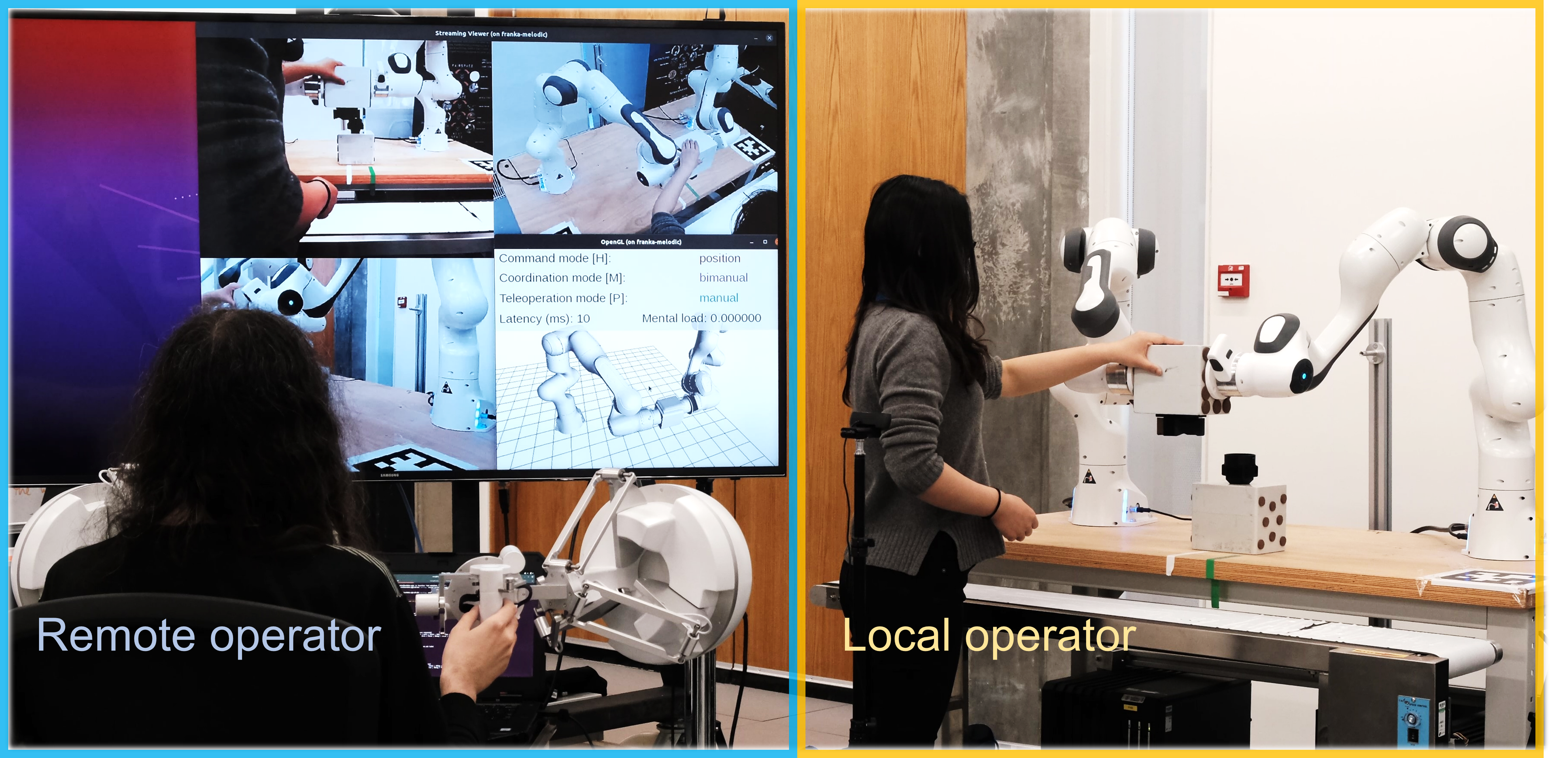}	
    \caption{Real-world multi-operator part assembly task by the remote and local operators using the proposed collaborative bimanual manipulation framework.}
    \label{fig:scenario}
    \vspace{-5mm}
\end{figure}

Teleoperation has great potential in bimanual manipulation tasks where human operators use master devices to control the robots remotely \cite{selvaggio2021shared}. Recent development in collaborative robots (CoBot) has also enabled the safe operation of the robots in close proximity to humans \cite{industry}. Human operators have complementary expertise which can be integrated into robotic control to increase the intelligence of the robotic system. Fig. \ref{fig:scenario} illustrates a multi-operator assembly task, where remote and local operators collaborate to control a dual-arm robotic system to assemble two parts. The remote operator has a global view of the situation, and teleoperates the dual-arm to reach, maintain contact with, hold and move the parts. The local operator can interactively guide and adjust the fine manipulation process using more detailed contact information. In both teleoperated and physically guided motions, human commands are subject to errors inevitably due to the lack of direct access and intuitive perception of robot's physical limits, which could result in the violation of the robots' physical constraints and cause task failures and safety issues. Therefore, the robot local control system should be robust to the infeasible commands for safe HRC, making it a necessity to adapt these references to feasible motions which can satisfy the physical limitations of the robot and the task constraints. 

In this work, we develop a collaborative bimanual manipulation framework for reliable and safe human-robot collaboration. For compliant physical interaction with local human operators, the operator's commands are estimated as external force which is sent to an admittance controller to modulate the contact forces between the object and two end-effectors. We propose an optimal motion adaptation framework--Task-Space Sequential Equilibrium and Inverse Kinematics Optimization (SEIKO) \cite{seiko} to adapt error-prone Cartesian human commands from physical interaction and teleoperation to feasible desired motions to satisfy the payload constraints, joint torque limits, and kinematic limits of the robots. The optimal and feasible commands generated by the Task-Space SEIKO are then realized by an interaction controller adopted from the Fractal Impedance Controller (FIC) scheme \cite{FIC2019} which superimposes multiple control efforts to generate desired joint torque commands to control dual-arm robots. FIC has a conservative observer which enables the stable superimposition of multiple controllers and is robust to time delay and the reduced control bandwidth. It enables the development of teleoperation setups with independent stability between the master and replica systems, and the admittance behavior on top of a soft-impedance controller \cite{Tiseo2021HapFic,tiseo2020}.

The proposed framework provides a viable solution for safe collaborative bimanual manipulation. The compatible integration of the admittance controller with the Task-Space SEIKO enables the successful manipulation of different objects without the need for tuning the controllers. Since SEIKO relies on a simplified dynamic model, only the mass, the center of mass (CoM) and the friction coefficient of the object need to be updated for different tasks. Meanwhile, we also consider the trade-off between admittance and impedance behaviors to reduce the Sim-To-Real gap due to the model inaccuracy.

The contributions of our work are summarized as follows:
\vspace{-2mm}
\begin{enumerate}

    \item \textbf{Task-Space SEIKO} that adapts infeasible Cartesian commands from operators to optimally satisfy the physical limits of robots and task constraints in real time (\SI{1}{\kilo\hertz});
    
    \item \textbf{Collaborative bimanual manipulation framework} that allows remote and local human operators to interactively control a dual-arm robotic system. A proposed interaction controller that adopted Fractal Impedance Control to generate coordinated motions and provide robust stability during human-robot interaction at a low computation cost;
    
    \item \textbf{Reliable collaborative manipulation} for bimanual tasks, which enabled arbitrary maneuver of a heavy object, reliable physical interactions with a stack of unknown objects, collaborative multi-operator part assembly, and industrial connector insertion.
\end{enumerate}

Our method has demonstrated the real-world performance in collaborative bimanual manipulation and safety guarantees for robots, the operating payload and human operators. It also provides the advantage of better decision making by fusing commands from multiple intelligent agents (human operators and robots) who specialize in different sub-tasks, thus creates opportunities for the explorations towards multi-agent intelligence, which will lower the physical and cognitive demands for human workers while improving the productivity and efficiency.

The paper is organized as follows. Section \ref{sec:biblio} summarizes the related works on human-robot collaborative bimanual manipulation. Section \ref{sec:method} overviews the control framework of the collaborative bimanual manipulation and formulates the Task-Space SEIKO and the interaction controller. Section \ref{sec:results} details the experimental setups and evaluates the performance of the proposed method, and Section \ref{sec:conclusions} draws the conclusions.

\section{Related Works}\label{sec:biblio}

%
Inverse kinematics (IK) has been extensively used to compute whole-body joint positions to get the desired end-effectors' poses while considering kinematic constraints in teleoperation. For example, the quasi-static equilibrium of humanoids on flat ground is formulated as an IK problem where the CoM projection is constrained within the support polygon. However, the contact wrenches which are important for bimanual manipulation tasks, are not considered in IK-based schemes. For loco-manipulation tasks in multi-contact settings, a motion retargeting framework -- SEIKO is proposed in \cite{seiko} and has been validated by the teleoperation of high DoF robots such as humanoids and quadrupeds. The retargeting of multi-contact motions is formulated as sequential quadratic programming which can optimize the whole-body configurations in real time. Although designed for the contact switch in multi-contact teleoperation tasks with quasi-static motions, SEIKO has great potential in the motion adaptation for bimanual manipulation tasks by adding the constraints of the contact wrenches in the static equilibrium equation of the object.

Direct human-robot collaboration requires physical interaction between humans and robots, either by direct contact or indirect force exchange through an object. The applications of physical human-robot interaction range from collaborative manufacturing such as assembly \cite{CHERUBINI20161} to emotional support such as hugging robots \cite{hugging}. Impedance control \cite{hogan2018impedance}. Admittance control \cite{admittance} are typically used for physical human-robot interaction to realize the desired dynamic behavior at its ports of interaction with the environment and their improved control for reactive interaction has enabled the close and efficient collaboration between robots and humans.

Multiple admittance and impedance controllers have been proposed to modulate the contact port performances to adapt to different task requirements over the years by exploiting both optimization and learning methods \cite{ajoudani2017choosing,khoramshahi2019dynamical,rakita2019shared}. Although it is theoretically possible to combine the admittance controller with any passive impedance controller, it has limited responsiveness and tracking accuracy. FIC controller has the advantages of a passive impedance controller while improves the responsiveness and tracking accuracy \cite{Tiseo2021HapFic}. Its highly non-linear intrinsically stable dynamics allows it to track trajectories accurately, while retaining the intrinsically soft behavior of impedance controllers \cite{FIC2019}. FIC can be combined with admittance controllers because it allows the superimposition of multiple independent controllers while retaining the system stability. 

However, the ability to generate robust interaction with high tracking accuracy is not sufficient to guarantee the success of bimanual manipulation tasks. The challenges of these tasks include the coordination of the two arms, and the retargeting of human commands that violate the robots' physical limits \cite{mansard2009stackoftasks,hutter2014excitation,dehio2020enabling,yan2021decentralized}. The coordination of multiple robotic arms requires the formulation of dynamic-aware motion optimization that deals with the intrinsic non-linearity and the non-holonomic constraints in concave domains. There are multiple ways to deal with these constraints from domain segmentation in sequences of overlapping convex sub-domains to maximizing the feasibility probability, such as Hierarchical Quadratic Programming (H-QP)\cite{mansard2009stackoftasks,hutter2014excitation,dehio2020enabling} and Non-Linear Programming (NLP)\cite{yan2021decentralized}.
\begin{figure*}[t]
    \begin{subfigure}{1\linewidth}
        \centering
        \includegraphics[trim=0cm 0cm 0cm 0cm,clip, width=\textwidth]{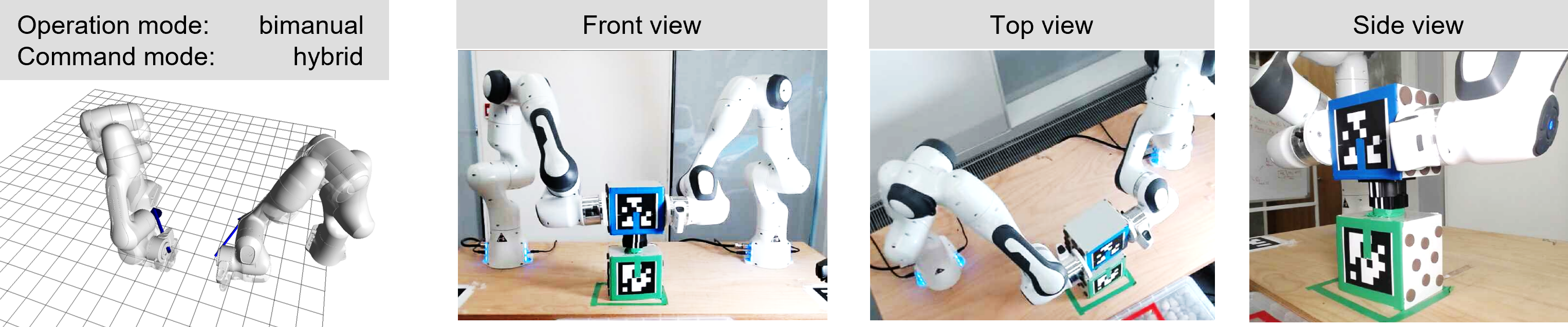}
        \caption{The teleoperation interface with multi-camera views. The interactive 3D scene is rendered from the robot state for the remote operator to switch mode. The images show that the remote user is operating in the bimanual mode with hybrid command mode.}
    \label{fig:teleop_interface}
    \end{subfigure}
    
    \begin{subfigure}{1\linewidth}
        \centering
        \includegraphics[trim=0.5cm 2cm 0cm 0cm,clip, width=\textwidth]{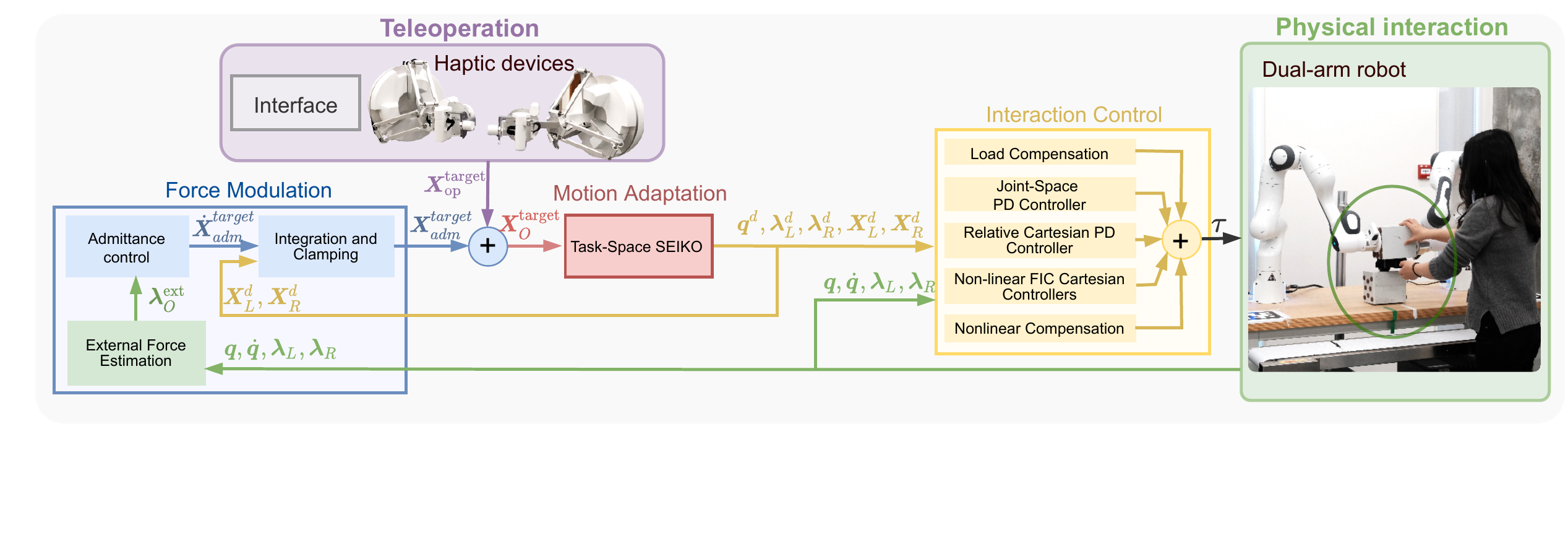}
        \caption{The operators' commands from the teleoperation interface and the external force feedback estimated from the physical interaction are combined to generate the desired object trajectory. The Task-Space SEIKO adapts this trajectory to satisfy the physical limits of the robotic system and the task constraints. The interaction controller implements the superimposed controllers including the passive impedance controllers (Joint-Space PD, Cartesian Coordination, and Non-Linear FIC) to guarantee the stability during human-robot interaction, and the compensatory controllers for the load compensation and the non-linear dynamics (Gravity, Coriolis and centrifugal effects).
    }
    \label{fig:control_diagram}
    \end{subfigure}
    \caption{The overall architecture of the collaborative bimanual manipulation framework.}
    \label{fig:architecture}
    \vspace{-3mm}
\end{figure*}

The H-QP solves this problem by dividing the action into a hierarchical set of tasks. The decomposition of the problem into sub-problems simplifies the constraints by decoupling them, while regarding the co-domain of the higher task as the domain of the current optimization \cite{mansard2009stackoftasks,hutter2014excitation}. NLP uses a general formulation to tackle non-linear optimizations \cite{yan2021decentralized}. H-QP and NLP can be formulated both in inverse and forward dynamics\cite{mansard2009stackoftasks}. It shows that the formulation of the inverse dynamics has faster convergence when using simultaneous methods (e.g., direct transcription) for NLP. These methods decompose the problem into a discrete set of sub-problems enforcing boundary constraints during the transitions between subsequent domains, which guarantees a smooth transition. The parallel processing of the optimization problem solves complex mathematical problems faster, which allows modeling the mechanism capabilities more accurately.

These methods have become increasingly more efficient over the last decade, and the computational time have been reduced drastically. Nevertheless, the coordinated dynamic interaction of multiple robotic arms is still an open problem in robotics, due to the intrinsic variability in the interaction with the  environment that is either unknown or difficult to model. The gap between simulation and reality will often reduce the robustness of interaction. 

\section{Collaborative Manipulation Framework}\label{sec:method}

The control framework of the collaborative bimanual manipulation is shown in Fig. \ref{fig:architecture}. The human commands can be given via the teleoperation interface or by physical interaction. The proposed framework consists of three parts: the admittance controller for compliant physical interaction and modulating the contact forces, the Task-Space SEIKO that optimizes the target pose of the object $\bm{X}_O^\text{target} \in \SE$ subject to the contact constraints and the physical limits of the robot, and the interaction controller that generates the joint torque commands for robots' stability during the interaction.

We propose a task-space formulation of SEIKO for the motion adaptation of infeasible human commands. It runs one iteration per control cycle to optimize the feasible state for the next time step, with the computation time of less than \SI{1}{\milli\second}. The optimal commands generated by the Task-Space SEIKO are realized by the interaction controller, which has three FIC controllers and two compensatory terms to generate the desired joint torques for each arm. 
\subsection{Operator Interface}
We use three cameras as multi-camera feeds to capture live streams of the robot for the remote operators, and the interface is shown in Fig. \ref{fig:teleop_interface}. The remote operators can teleoperate the robot via two 6-DoF haptic devices (Force Dimension Sigma 7) using two operation modes: the independent and the bimanual mode. We also design two command modes: twist or pose mode to be combined with these two operation modes. The pose command mode uses the displacement of the current pose from the nominal pose measured by the Sigma 7 haptic device as the control reference of the robot, which is more intuitive to operators. The twist mode has better use of the limited workspace of the haptic device as it can accumulate the displacement of the operator's pose from the nominal pose as the desired end-effector pose. In the independent mode, the remote operator teleoperates the left end-effector via the left haptic device and the right end-effector via the right device in the Cartesian space, using either command mode (pose/twist) or a hybrid of both. After placing the object between the two hands, the operator can enable the bimanual mode in which the human commands are sent to control the pose of the CoM of the object via one haptic device. In bimanual mode, the target pose commands from the operator are optimized by Task-Space SEIKO to satisfy the task constraints and physical limits of the robots.

The controller used for the haptic feedback on the Sigma 7 is adopted from our previous work and its technical details can be found in \cite{tiseo2021fine}. The target end-effector pose $\bm{X}_\text{op}^\text{target}$ for the two command modes in the bimanual mode can be computed by Eq. (\ref{eq:operator}) from human commands, which will be sent to the Task-Space SEIKO for motion adaptation.
\begin{equation}
\label{eq:operator}
\begin{array}{lc}
    \bm{X}_\text{op}^\text{target} = \bm{X}_\text{ref}^\text{target} + \Delta\bm{X}_\text{S-7}\text{,~~~ pose mode}\\
    \bm{X}_\text{ref}^\text{target} = \bm{X}_{\text{ref}0}^\text{target} + \Delta\bm{X}_\text{S-7}\Delta t\text{,~~~ twist mode}
    \end{array}
\end{equation}
where $\bm{X}_\text{ref}^\text{target}$ is the reference rest pose of the end-effector, $\Delta\bm{X}_\text{S-7}$ is the displacement of the current pose from the nominal pose measured by the Sigma 7 haptic device, and $\bm{X}_{\text{ref}0}^\text{target}$ is the end-effector's reference pose of the arm at the last time step.

\begin{figure}[t]
    \centering
	\includegraphics[trim=1cm 0 0.6cm 0.6cm,clip,width=0.8\linewidth]{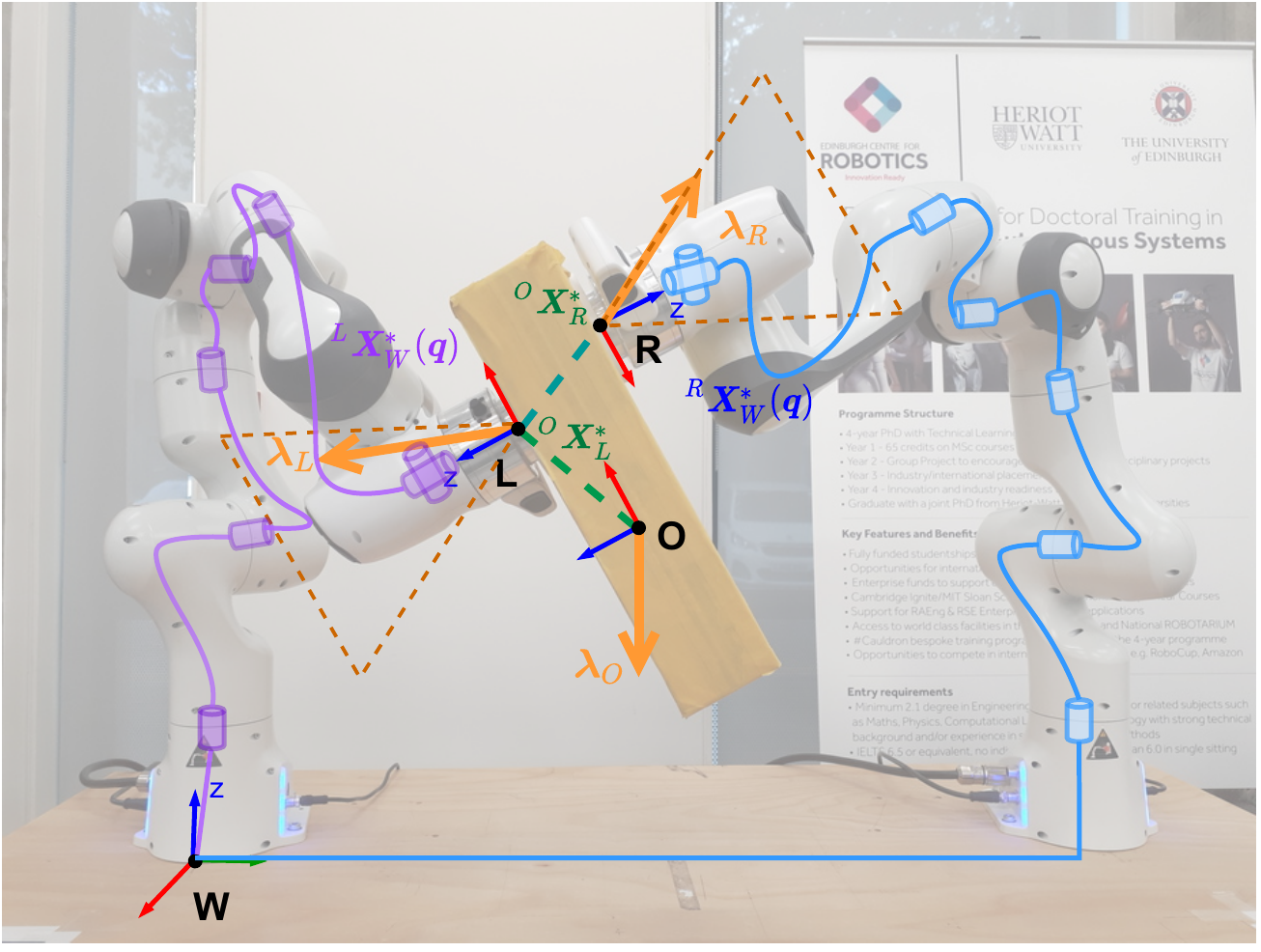}	
    \caption{
    The illustration of the reference frames, the force and spatial transform notations used in the optimal motion adaptation. The end-effector frames of the left hand ($L$) and the right hand ($R$), as well as the object frame (the origin $O$ is located at its CoM) are depicted. The world frame $W$ is placed at the base of left arm. The contact wrenches between the object and the left and right arms are denoted by $\bm{\lambda}_L, \bm{\lambda}_R$, respectively, while $\bm{\lambda}_O$ denotes the gravitational wrench. The friction cones of the two contact wrenches are depicted by the orange dash lines.
    }
    \label{fig:object_schema}
    \vspace{-5mm}
\end{figure}
\subsection{End-Effector Force Modulation}\label{adm-control}
The model-based state estimation uses an admittance controller to track the desired interaction force at the end-effector, which takes the force tracking error as input and generates the end-effectors' pose commands. The admittance controller and the Task-Space SEIKO handles the task performance independently and drive the impedance controller to generate the desired behavior. The admittance controller filters the measured force tracking error and generates the desired robot end-effector pose that can be adaptive to different tasks by producing a desired interaction dynamics at the end-effector, as follows: 

\begin{equation}
\label{eq:TaskStateEstimator}
    \bm{\ddot{X}}_{adm}^{target}=\bm{M}^{-1}\left(\bm{\lambda}_O^{ext}-\bm{\lambda}^{d}_\text{adm}\right),
\end{equation}
where $\bm{M}^{-1}$ is the desired inertia at the end-effector, $\bm{\lambda}_O^{ext}$ is the estimated end-effector effort, and the $\bm{\lambda}^{d}_\text{adm}$ is the desired interaction wrench. $\bm{\lambda}^{d}_\text{adm}$ can be treated either as an external input or modeled as a pre-assigned spring-damper system as follows:
\begin{equation}
\label{eq:TaskStateEstimatorSD}
    \bm{\lambda}^{d}_\text{adm}=\bm{K}_d\left(\bm{X}^d-\bm{X}\right)-\bm{D}_d\bm{\dot{X}},
\end{equation} 
where $\bm{K}_d$ and $\bm{D}_d$ are the desired damping and stiffness, respectively. The desired acceleration of the two arms is integrated twice to get the desired end-effector pose ($\bm{X}_{adm}^{target}$), as follows:
\begin{equation}
\label{eq:TaskStateIntegration}
    \begin{cases}
       \bm{\dot{X}}_{adm}^{target}(t+\Delta t)=\bm{\dot{X}}_{adm}^{target}(t) + \Delta\bm{\dot{X}}_{adm}^{max}(t)\\\\
       \bm{X}_{adm}^{target}(t+\Delta t)=\bm{X}_{adm}^{target}(t) + \bm{\dot{X}}_{adm}(t) \Delta t
    \end{cases},
\end{equation}
\begin{equation*}
\begin{array}{l}
    \text{where ~}  \\
    \Delta\bm{\dot{X}}_{adm}^{max}=\mathsf{sign}\left(\bm{\ddot{X}}_{adm}\right)\mathsf{max}\left(\left|\bm{\ddot{X}}_{adm}\right|\Delta t,\Delta \bm{\dot{X}}_{max}\right)
    \end{array},
\end{equation*}
where $\Delta \bm{\dot{X}}_{max}=|| \bm{X}_{O}^{target}(t)-\bm{X}_{O}^{d}(t)||/\Delta t$ is the feasibly maximum twist change of the selected arm at the current time step, bounding the end-effector twist to the feasibility sphere around the current end-effector state effectively . 

The state estimator in Fig. \ref{fig:control_diagram} estimates the external interaction with the manipulated object by subtracting the weight of the held object from the wrench computed from the superimposition of the measured wrenches at the robot end-effectors $\bm{\lambda}_O^{ext}={}^O\bm{X}_{L}^*\bm{\lambda}_{L}+{}^O\bm{X}_{R}^*\bm{\lambda}_{R}$, where ${}^O\bm{X}_{L}^*$ and ${}^O\bm{X}_{R}^* \in \R^{6 \times 6}$ are the wrench transformation matrix between the robot end-effector frames of the two arms ($L,R$) and the object frame ($O$). $\bm{\lambda}_{L}$ and $\bm{\lambda}_{R} \in \R^6$ are the measured end-effector wrenches for the left and right arm, respectively.

\subsection{Motion Adaptation}
\label{sec:SEIKO}
The motion adaptation method takes the end-effector pose commands as input and solves the posture and contact forces $(\bm{q},\bm{\lambda})$ simultaneously as a nonlinear optimization problem. We formulate a Task-Space SEIKO in which the equilibrium of the external object held between the two arms is expressed in the Cartesian space. The reference frames used to estimate the object's static equilibrium from the robot state is shown in Fig. \ref{fig:object_schema}. SEIKO transforms an input target pose of the CoM of the manipulated object $\bm{X}_O^{\text{target}} \in \SE$ expressed in world frame to the optimized state of the arms that guarantees feasibility. The Task-Space SEIKO executes the optimization one iteration at each control cycle. At each iteration, the state change $(\Delta\bm{q},\Delta\bm{\lambda})$ is computed to update the previous state to a new desired state $\bm{q}^d \in \R^n$, where $n$ is the total number of the left and right robotic arms, and the desired wrenches applied on the object $\bm{\lambda}_L^d, \bm{\lambda}_R^d \in \R^6$ by the two arms. The Cartesian poses $\bm{X}_L^{\text{d}}, \bm{X}_R^{d} \in \SE$ of the two end-effectors are then computed by forward kinematics.

The task-space SEIKO is formulated as a constrained nonlinear optimization and solved by a sequence of QP problems. Both the cost function and the constraints are linearized and approximated at the first-order and rely on analytical derivatives for computational speed and stability. The constrained least square optimization problem to be solved at each control loop is as follows:
\begin{equation}
\begin{aligned}
    \label{eq:nonlinear}
     \underset{\Delta\bm{x}}{\text{min}}~\norm{\bm{C}_{\text{cost}}(\bm{x}) \Delta\bm{x} -  \bm{c}_{\text{cost}}(\bm{x})}^2_{\bm{w}} \\
    \text{s.t.~} \bm{C}_{\text{eq}}(\bm{x}) \Delta\bm{x} + \bm{c}_{\text{eq}}(\bm{x}) &= \bm{0}, \\
                \bm{C}_{\text{ineq}}(\bm{x}) \Delta\bm{x} + \bm{c}_{\text{ineq}}(\bm{x}) &\geqslant \bm{0}, \\
    \text{where } \bm{x} = \mat{\bm{q}^d \\ \bm{\lambda}^d_L \\ \bm{\lambda}^d_R},
    \Delta\bm{x} &= \mat{\Delta\bm{q} \\ \Delta\bm{\lambda}_L \\ \Delta\bm{\lambda}_R}, \\
\end{aligned}
\end{equation}
where $\bm{x} \in \R^{n+6+6}$ is the current desired state, and the incremental change $\Delta\bm{x}$ is the decision variable. $\bm{q}^d \in \R^{n}$ is the vector containing all joint positions of the two robotic arms, $\bm{C}_{\text{cost}},\bm{c}_{\text{cost}},\bm{C}_{\text{eq}},\bm{c}_{\text{eq}},\bm{C}_{\text{ineq}},\bm{c}_{\text{ineq}}$ are the matrices and vectors defining the cost, equality and inequality constraints respectively. Our decision variable here is the incremental change $\Delta\bm{x}$, and the desired state is updated at each iteration by $\bm{x}(t+\Delta t) = \bm{x}(t) + \Delta\bm{x}(t)$.
The equality and inequality constraints are detailed as follows.

\subsubsection{Equality constraints}
To adapt the motions within the stability region for the success of bimanual manipulation tasks, we introduce a constraint term based on the grasp matrix and the robot state to the given optimization formulation. The frames used to formulate the optimization problem include the world frame $\bm{W}$, the object frame $\bm{O}$, as well as the left and right end-effector frame $\bm{L}$, $\bm{R}$, respectively, depicted in Fig.~\ref{fig:object_schema}. We align the object frame $\bm{O}$ to the frame $\bm{L}$ and set its origin at the CoM of the object for a simplified static equilibrium equation. The position of $\bm{O}$ can be estimated from the contact state.

The static equilibrium equation of an object in the bimanual manipulation task is as follows:
\begin{equation}
    \label{eq:static_constraint}
    {}^O\bm{X}_W^*(\bm{q}) \ \bm{\lambda}_{\text{O}} = [{}^O\bm{X}_L^* \ {}^O\bm{X}_R^*]
    \mat{\bm{\lambda}_L\\ \bm{\lambda}_R},
\end{equation}
where $\bm{\lambda}_L, \bm{\lambda}_R \in \R^6$ are the wrenches expressed in the left and right end-effector frame $\bm{L}$, $\bm{R}$, while ${}^O\bm{X}_{L}^*$ and ${}^O\bm{X}_{R}^* \in \R^{6 \times 6}$ are the spatial wrench transformation matrix (\cite{featherstone2014rigid}) from $\bm{L}$ and $\bm{R}$ to the object frame $\bm{O}$, respectively. The $Z$ axis of $\bm{L}$ and $\bm{R}$ are in parallel to the normal direction of the contact surface. $\bm{\lambda}_{\text{O}} = [\bm{\tau}_{\text{O}} ~ \bm{f}_{\text{O}}]^\T=[0 ~ 0 ~ 0 ~ 0 ~ 0 ~ -mg]^\T$ is the gravitational wrench of the object expressed in the world frame $\bm{W}$, where $m$ is the object's mass and $g$ the gravitational acceleration.
${}^O\bm{X}_W^*(\bm{q}) \in \R^{6 \times 6}$ is the wrench transformation from the world frame $\bm{W}$ to the object frame $\bm{O}$.
Because $\bm{\lambda}_{\text{O}}$ does not have a torque component ($\bm{\tau}_{\text{O}} = \bm{0}$) since it acts on the CoM of the object, ${}^O\bm{X}_W^*(\bm{q})$ can be simplified as $\mat{\bm{0} & \bm{0} \\ \bm{0} & {}^O\bm{R}_W(\bm{q})}$, where
${}^O\bm{R}_W(\bm{q}) \in \SO$ is the rotational term of ${}^O\bm{X}_W^*(\bm{q})$, which only depends on the joint positions. 
 
The equality constraint for static equilibrium in Eq. (\ref{eq:static_constraint}) is differentiated with respect to $[\Delta\bm{q} ~ \Delta\bm{\lambda}_L ~ \Delta\bm{\lambda}_R]^\T$, resulting in the following equation:

\begin{equation}
\label{eq:diff_static_constraint}
\begin{aligned}
    & \mat{\mat{\bm{0} \\ \bm{H}(\bm{q}, \bm{f}_{\text{O}})} & {}^O\bm{X}_L^* & {}^O\bm{X}_R^*}\mat{\Delta\bm{q} \\ \Delta\bm{\lambda}_L ~\\ \Delta\bm{\lambda}_R} + \\
    &~~~~~~~~ {}^O\bm{X}_W^*(\bm{q})\bm{\lambda}_{\text{O}} - {}^O\bm{X}_L^*\bm{\lambda}_L + {}^O\bm{X}_R^*\bm{\lambda}_R = \bm{0}, \\
    & \text{where ~} \bm{H}(\bm{q}, \bm{f}_{\text{O}}) \in \R^{3 \times n} = \\
    &~~~~~~~~ -\frac{1}{2}{}^O\bm{R}_L\bm{S}\left({}^L\bm{R}_W(\bm{q})\bm{f}_{\text{O}}\right){}^L\bm{J}_W^{\text{rot}}(\bm{q}) \\
    &~~~~~~~~ -\frac{1}{2}{}^O\bm{R}_R\bm{S}\left({}^R\bm{R}_W(\bm{q})\bm{f}_{\text{O}}\right){}^R\bm{J}_W^{\text{rot}}(\bm{q}), \\
\end{aligned}
\end{equation}
and where ${}^L\bm{J}_W^{\text{rot}}(\bm{q}), {}^R\bm{J}_W^{\text{rot}}(\bm{q}) \in \R^{3 \times n}$ are the angular parts of the Jacobians of the frames $\bm{L}$ and $\bm{R}$ expressed in world frame $\bm{W}$, respectively.
${}^L\bm{R}_W(\bm{q}), {}^R\bm{R}_W(\bm{q}), {}^O\bm{R}_L(\bm{q}), {}^O\bm{R}_R(\bm{q}) \in \SO$ are the spatial rotational transformations between the frames depicted in Fig.~\ref{fig:object_schema}.
$\bm{S}(\cdot): \R^3 \to \R^{3 \times 3}$ is the operator generating the skew matrix associated to a 3d vector.
We rearranged the terms in $\bm{H}(\bm{q}, \bm{f}_{\text{O}})$ so that Eq. (\ref{eq:diff_static_constraint}) is linear with respect to the decision variable $\Delta \bm{x}$ and is suitable for the QP formulation $\bm{C}_{\text{eq}}$ and $\bm{c}_{\text{eq}}$ in Eq. (\ref{eq:nonlinear}).

\subsubsection{Inequality constraints}
The inequality constraints used in the optimization enforce the feasibility of the bimanual manipulation. Two sets of different constraints are defined. In joint space, we enforce the physical limits of the robotic arms by setting constraint for each joint: the maximum torque ($\bm{\tau}^d$), the maximum velocity ($\bm{\dot{q}}^d$) and the angular position range ($\bm{q}^d$). In task space, we constrain the contact wrenches ($\bm{\lambda}_L^d, \bm{\lambda}_R^d$) to enforce the stability of the contact, e.g. minimal and maximal normal contact force and limits on friction pyramid and center of pressure. All these constraints can be written in the linear form $\bm{C}_{\text{ineq}}, \bm{c}_{\text{ineq}}$ with respect to the decision variable $\Delta \bm{x}$ (see \cite{seiko}), spawning $4n$ rows for the joint space and $36$ rows for the task space contact.

\vspace{-10mm}

\subsection{Interaction Controller}
The interaction controller adopted from FIC, which is robust to time delay and stable superimpose multiple control efforts, is used to generate desired joint torques and control the dual-arm robot. As shown in Fig.\ref{fig:control_diagram}, the interactive controller is composed of five independent controllers. 

The non-linear FIC Cartesian controller ($\mathsf{NLPD}(\bm{X}^d, \bm{X}, \bm{\nu})$) independently drives the arms' end-effectors towards their respective desired poses. The linear task-space PD controller ($\mathsf{PD}(\bm{X}^d_{\text{rel}}, \bm{X}_{\text{rel}}, \bm{\nu}_{\text{rel}})$) generates a wrench command to maintain the desired relative pose between the two arms. The linear joint-space PD controller  ($\mathsf{PD}(\bm{q}^d, \bm{q}, \bmd{q})$) drives the arms towards the desired poses optimized by the Task-Space SEIKO. 
\begin{figure*}[t]
    \centering
    \begin{subfigure}[b]{\textwidth}
    \centering
		\includegraphics[trim=0 0cm 0 0cm,clip,width=\textwidth]{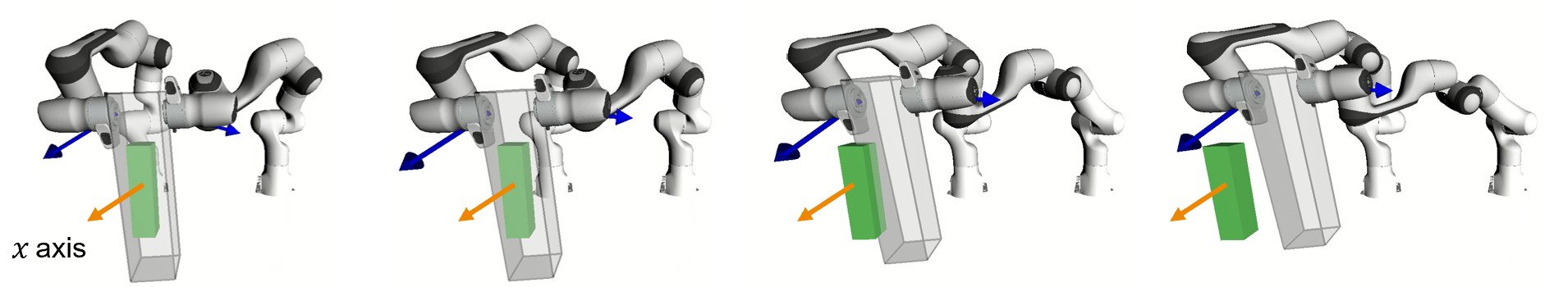}	
		\caption{}
		\label{fig:sim_trans}
	\end{subfigure}
	
	\begin{subfigure}[b]{\textwidth}
	\centering
		\label{fig:vio_success}
		\includegraphics[trim=0 0cm 0 0cm,clip,width=\textwidth]{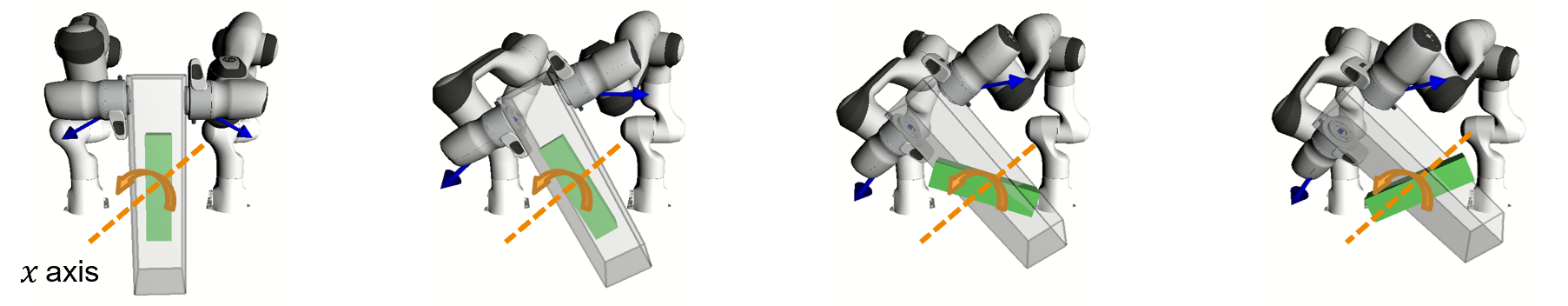}	
		\caption{}
		\label{fig:sim_rotate}
	\end{subfigure}

    \caption{Validation of feasibility adaptation visualized by the original references (green boxes) of arbitrary human commands versus the optimized adaptations (transparent boxes) with contract forces (blue arrows) in (a) translational and (b) rotational tasks respectively. The orange arrows indicate the directions of the translational and rotational motions.}
    \label{fig:visual}
    \vspace{-3mm}
\end{figure*}

\begin{table}[t]
    \centering
    \caption{Control parameters for the joint space, and linear and angular control in Cartesian space.}
    \label{table:control_parameters}
    \begin{tabular}{l c c l c c}
        \hline
        \textbf{Parameter} & \textbf{Value} & \textbf{Unit} & \textbf{Parameter} & \textbf{Value} & \textbf{Unit}\\
        \hline
        $f^{\text{lin}}$ & $10$ -- $40$ &$\SI{}{\newton}$ & $f^{\text{ang}}$ & $2.0$ & $\SI{}{\newton\meter}$\\
        $d^{\text{lin}}$ & $0.08$ & $\SI{}{\meter}$&$d^{\text{ang}}$ & $8.0$ & degree\\
        $\zeta^{\text{lin}}$ & $0.8$ & & $\zeta^{\text{ang}}$ & $0.2$ &\\
        \hline        
        $f^{\text{rel,lin}}$ & $50.0$ & $\SI{}{\newton}$&$f^{\text{rel,ang}}$ & $5.0$ & $\SI{}{\newton}$\\
        $d^{\text{rel,lin}}$ & $0.05$ & $\SI{}{\meter}$ & $d^{\text{rel,ang}}$ & $5.0$ & degree\\
        $\zeta^{\text{rel,lin}}$ & $0.4$ & & $\zeta^{\text{rel,ang}}$ & $0.1$ &\\
        \hline
        $f^{\text{joint}}$ & $0.3$ & $\SI{}{\newton\meter}$ & & &\\
        $d^{\text{joint}}$ & $10.0$ & degree & & &\\
        $\zeta^{\text{joint}}$ & $0.0$ & & & & \\
        \hline
    \end{tabular}
    \vspace{-5mm}
\end{table}

The remaining components of the controller compensate for the arm non-linear dynamics (Coriolis matrix and gravity compensation) and the load weight ($\bm{\lambda}^\text{d}$). The torque command sent to the robotic arms is as follows:
\begin{equation}
\begin{aligned}
    \bm{\tau} =~ & \bm{C}(\bm{q}, \bmd{q}) + \bm{G}(\bm{q}) ~+ \\
    & \mathsf{PD}(\bm{q}^d, \bm{q}, \bmd{q}) ~+\\
    & \bm{J}_{\text{local}}(\bm{q})^\T \bm{\lambda}^\text{d} ~+\\
    & \bm{J}_{\text{world}}(\bm{q})^\T \mathsf{NLPD}(\bm{X}^d, \bm{X}, \bm{\nu}) ~+\\
    & \bm{J}_{\text{rel}}(\bm{q})^\T \mathsf{PD}(\bm{X}^d_{\text{rel}}, \bm{X}_{\text{rel}}, \bm{\nu}_{\text{rel}})
\end{aligned},
\end{equation}
where $\bm{C(\bm{q}, \bmd{q})} \in \R^n$ is the vector containing Coriolis and centrifugal torques, while $\bm{G(\bm{q})} \in \R^n$ is the vector containing the gravitational torques;
$\bm{q} \in \R^{n}$ and $\bmd{q} \in \R^{n}$ are the vectors containing the measured joint positions and velocities of the two arms;
$\bm{q}^d \in \R^{n}$ is the desired joint positions computed by SEIKO;
$\bm{J}_{\text{local}}$ and $\bm{J}_{\text{world}} \in \R^{(6+6) \times n}$ are the stacked Jacobian matrices of the two arms expressed in local and world frames respectively, and $\bm{J}_{\text{rel}} \in \R^{6 \times n}$ is the relative Jacobian between the two hands;
$\bm{\lambda}^d \in \R^{6+6}$ is the stacked vector of the desired contact wrenches of two arms computed by SEIKO, applied as a feedforward term;
$\bm{X}$, $\bm{X}^d \in \SE\times\SE$ and $\bm{\nu} \in \R^{6+6}$ are the measured and desired Cartesian poses and measured twist of the two hands expressed in world frame; $\bm{X}_{\text{rel}}$ and $\bm{X}^d_{\text{rel}} \in \SE$ and $\bm{\nu}_{\text{rel}} \in \R^6$ are the measured and desired relative poses and measured relative twist between the two hands, respectively.

\begin{figure*}[t]
    \centering
    \begin{subfigure}{1\linewidth}
		\centering
		\includegraphics[trim=0 0 0 0,clip, width=17.6cm]{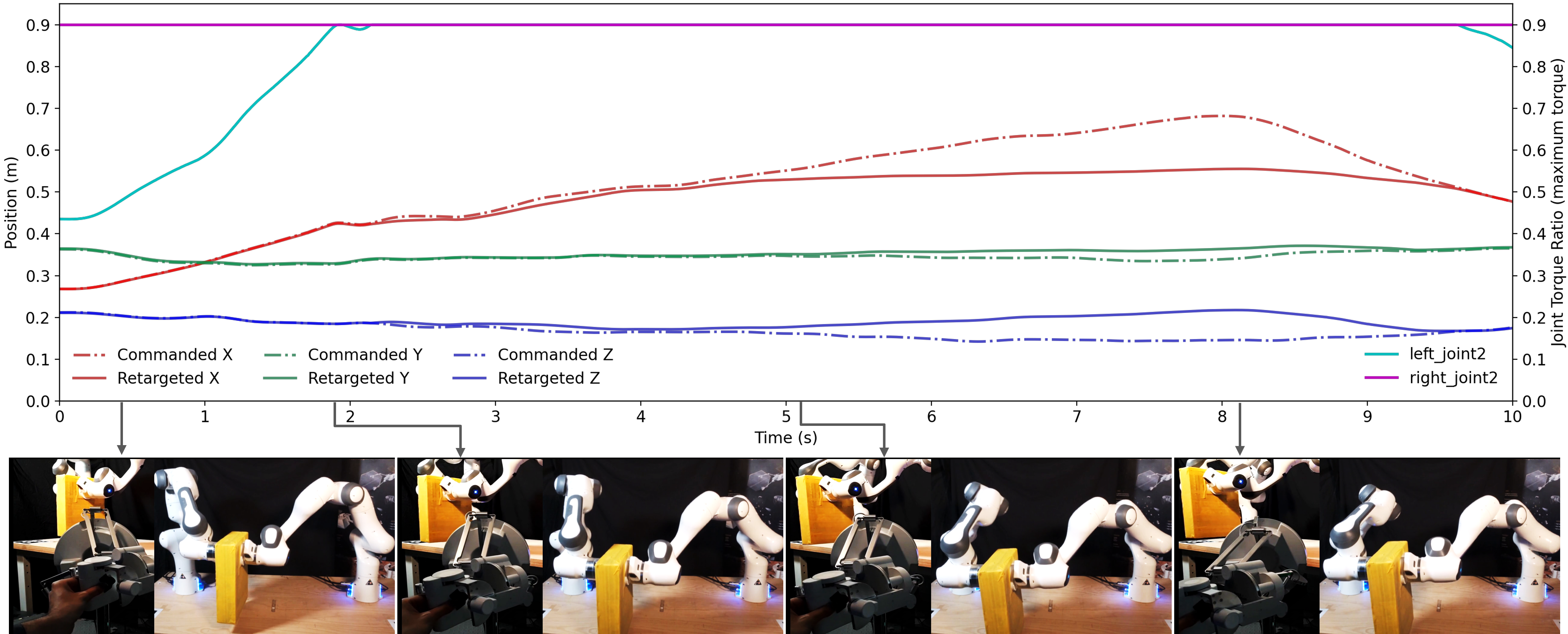}	
		\caption{The Cartesian positions of the CoM of the box in the world frame during the teleoperation of moving the box along the $x$ axis.}
		\label{fig:exp_box_trans}
	\end{subfigure}
	
	\begin{subfigure}{1\linewidth}
		\centering
		\includegraphics[trim=0 0 0 0,clip,width=17.6cm]{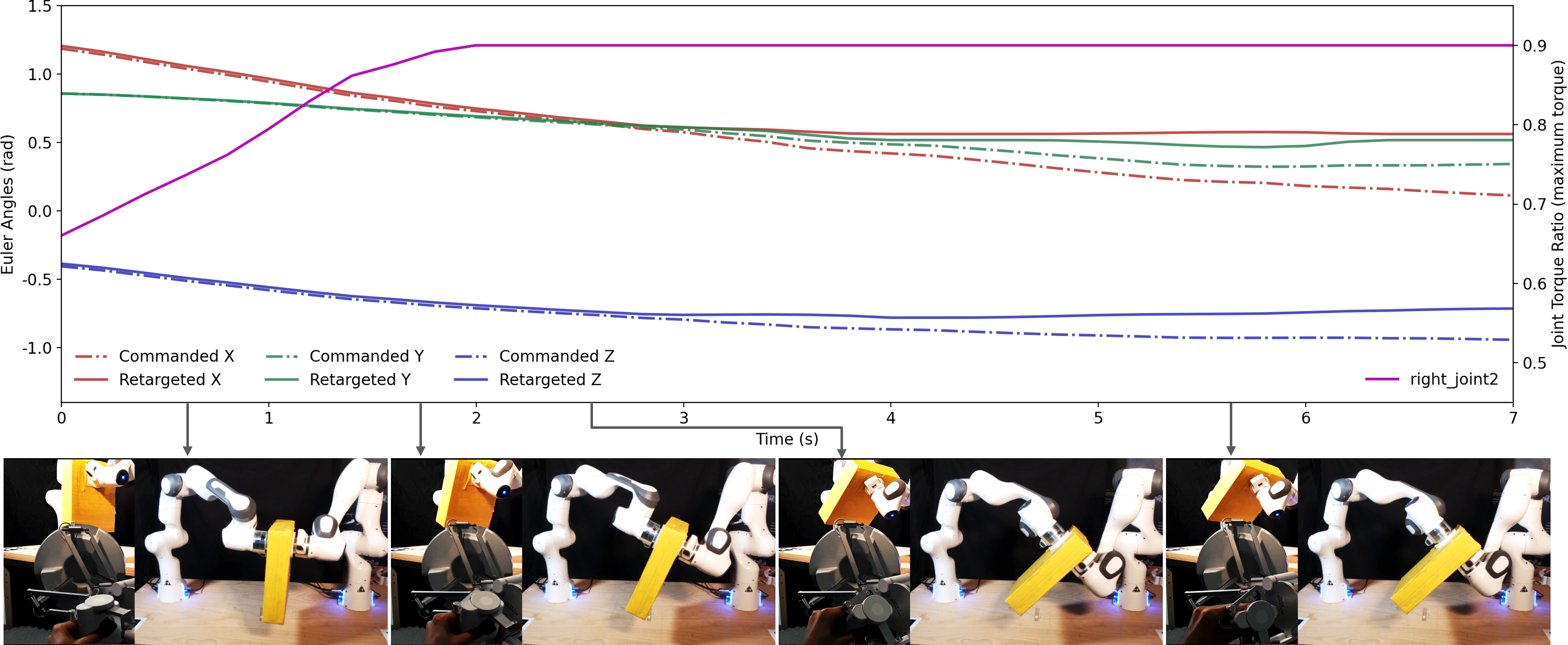}	
		\caption{The Cartesian orientations of the CoM of the box in the world frame during the teleoperation of rotating the box around the $x$ axis.}
		\label{fig:exp_box_rotate}
	\end{subfigure}

    \caption{
    The Cartesian trajectories of the CoM of the box along with the snapshots of the robotic arms and haptic devices during the teleoperation of translational and rotational motions. The data plots of the commanded (dashed lines) and the retargeted motions (solid lines) show that the proposed Task-Space SEIKO can adapt the human Cartesian commands to satisfy the joint torque limits.}
    \label{fig:exp_violation}
    \vspace{-5mm}
\end{figure*}

The passive PD controllers are implemented using six decoupled dimensions of the following mono-dimensional controller:
\begin{equation}
\begin{array}{l}
\mathsf{PD}(\alpha^d,\alpha, \dot{\alpha}) = F_k-k_d\dot{\alpha},\\
F_k=\begin{cases}
    k_p(\alpha^d-\alpha), & \text{if~} |\alpha^d-\alpha| < d \\
    \mathsf{sign}(\alpha^d-\alpha)f, & \text{else}
\end{cases}\end{array}
\end{equation}
where $\alpha$ is the position (i.e., mono-dimensional pose), $\alpha^d$ is the desired position, and $\dot{\alpha}$ is the velocity (i.e., mono-dimensional twist). The gains $k_p = f/d \in \R$ and $k_d^{\text{joint}} = 2\zeta\sqrt{k_p} \in \R$ which allows an intuitive tuning of the controllers.

The NLPD is implemented using six decouple dimensions of the following mono-dimensional controller:
\begin{equation}\label{eq:FICForce}
\begin{array}{l}
  \mathsf{NLPD}(\alpha^d,\alpha, \dot{\alpha}) = F_k-k_d\dot{\alpha},\\
  
    F_k=\left\{\begin{array}{ll}
        E(\tilde{\alpha}),  & \text{Divergence} \\
        \cfrac{2E(\tilde{\alpha}_\text{max})}{\tilde{\alpha}_\text{max}}\left(\tilde{\alpha}-\cfrac{\tilde{\alpha}_\text{max}}{2}\right) & \text{Convergence}
        \end{array}\right.\\
    E=\begin{cases}
        k_p\tilde{\alpha},& |\tilde{\alpha}|\le \xi d \\
        \cfrac{\Lambda}{2}\left(\tanh\left(\cfrac{\tilde{\alpha}-\tilde{\alpha}_\text{b}}{S\tilde{\alpha}_b}+\pi\right)+1\right)\\+E_0,& \text{else}
    \end{cases}
\end{array}
\end{equation}
where $k_p$ is the constant stiffness, $\tilde{\alpha}=\alpha_\text{d}-\alpha$ is the position error (i.e., mono-dimensional pose error), and $d$ is the tracking error when the force saturation occurs. $\Lambda=E_\text{max}-E_0$, $E_0=\xi k_p d$, $S=\left(1- \xi
\right)\left(d/(2\pi)\right)$ controls the saturation speed, and $\xi =0.9$ control the starting of the saturation behavior while approaching $d$.
Further details on the theoretical background of these controllers can be found in \cite{Tiseo2021HapFic}.
The parameters used for all the controllers are reported in Table \ref{table:control_parameters}.

\section{Experimental Results}\label{sec:results}
We set up two Franka Emika robotic arms as a dual-arm robot to validate the proposed collaborative bimanual manipulation framework with the following tasks. We first evaluated the motion adaptation ability of the proposed Task-Space SEIKO by teleoperating the robot to violate the joint torque limits. We then further validated the capability of the proposed method with collaborative bimanual manipulation tasks, including maneuvering a stack of objects via physical interaction, the multi-operator part assembly task and the industrial connector insertion.

In the joint torque limit violation experiment, we manually set the torque limits of both shoulder (joint 2) joints to a ratio (0.9) of their maximum torques to protect the robots from damage. An operator teleoperated the dual-arm robot to move an object with translational and rotational motions which could violate the joint torque limits of the robot. We first tested the proposed method in simulation and visualized the adapted motions of moving the box along $x$ axis and rotating the box around $x$ axis (in world frame) in Fig. \ref{fig:visual}. The green boxes indicate the object's original references calculated from arbitrary human commands, which were adapted to the feasible motions indicated by the transparent boxes.

We also evaluated our method on the real robotic system by teleoperating the dual-arm robot to violate the shoulder joint torque limits in a box moving task. The snapshots in Fig. \ref{fig:exp_box_trans} show the motions of the box when the remote operator moved the box along the $x+$ axis via two haptic devices, and the Cartesian trajectories of the CoM of the box include the human commands and the adapted motions. During the first 2 seconds, the right shoulder joint has reached its torque limit, while the joint torque of the left shoulder was still increasing. The Cartesian trajectories show that the robotic arms could still follow human commands because the proposed motion adaptation has the capability to optimize and adjust the joint-space configuration within the physical constraints of robots. After 2 seconds, the shoulder joints of both arms reached their joint torque limits, and the joint-space configuration cannot be further adjusted to follow human commands. The right two snapshots show that the robot stopped moving the box even though the remote operator was still giving commands to teleoperate the robot to move the box along the $x+$ direction. 

\begin{figure*}[t]
    \centering
    \includegraphics[trim=0cm 0cm 0cm 0cm,clip,width=0.95\textwidth]{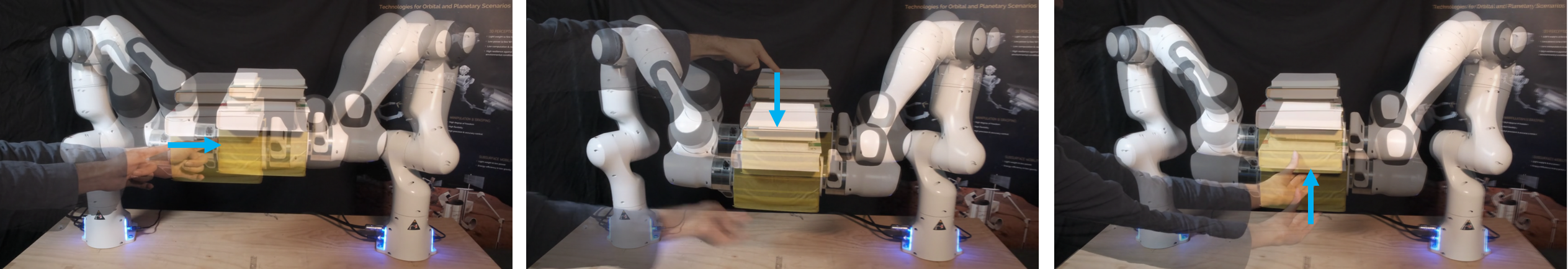}	
	\caption{A local operator and dual-arm robot collaborated to move a stack of books translationally via physical interaction.}
    \label{fig:exp_direct_books}
    \vspace{-5mm}
\end{figure*}

\begin{figure*}[t]
    \begin{subfigure}{1\linewidth}
    \centering	
	\includegraphics[width=0.95\textwidth]{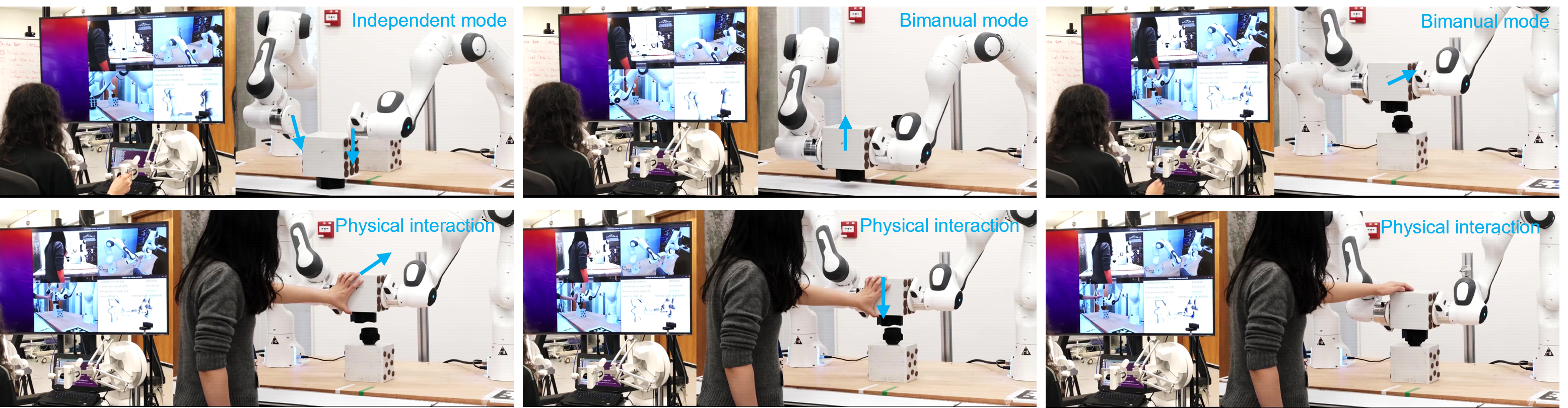}
    \caption{A multi-operator part assembly task}
    \label{fig:part_assembly}
    \end{subfigure}
    \vspace{5pt}
    \begin{subfigure}{1\linewidth}
    \centering	
	\includegraphics[width=0.95\textwidth]{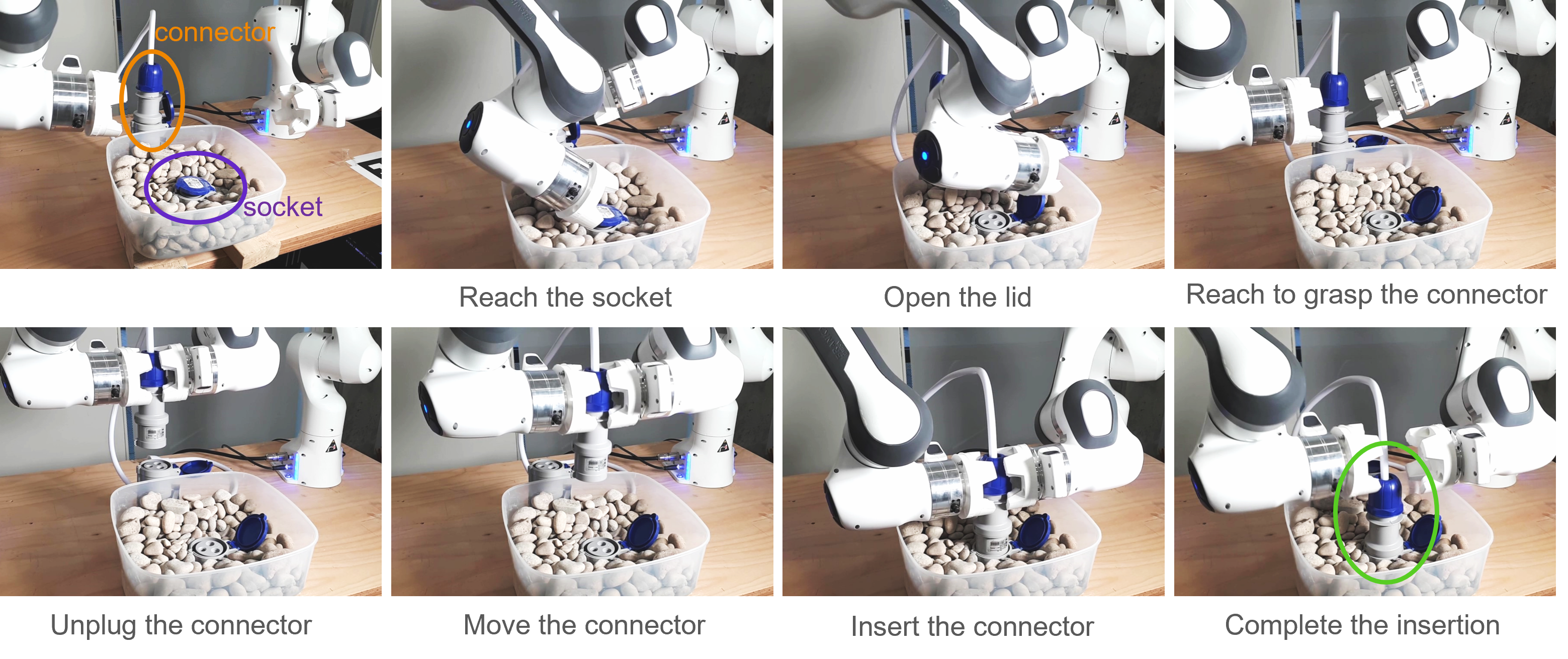}
    \caption{A teleoperated industrial connector insertion task}
    \label{fig:connector_insertion}
    \vspace{-2mm}
    \end{subfigure}
    \caption{Part assembly and connector insertion tasks using collaborative bimanual manipulation}
    \label{fig:exp_colab}
    \vspace{-5mm}
\end{figure*}

The process of teleoperating the robot to rotate the same box is shown in Fig. \ref{fig:exp_box_rotate}. We misaligned the CoM of the box with the center of the two end-effectors to create a large gravitational torque at each contact point, which will increase when rotating the box around the $x$ axis, making this task even more challenging. During the first 3 seconds, the robot could rotate the box to follow human commands, shown by the trajectories of Euler Angles in Fig. \ref{fig:exp_box_rotate}. However, the human commands were adapted after the right shoulder joint reached its torque limit. The right two snapshots show that the rotation of the box was stopped even though the human operator continued giving rotational commands through haptic devices. The success of this task validated that SEIKO is capable of the real-time motion adaptation by optimizing the maximum rotational motions of the end-effectors subject to joint torque limits and the static equilibrium by generating enough contact forces while keeping them within the friction cones. 

As comparison, the same experiments were conducted with the motion adaption method being disabled in the control framework. However, slippage and crashing happened during teleoperation, which can be seen in the accompanying video. The robotic arms failed to apply enough contact forces when the shoulder joint torque limits were violated. Results of the joint torque violation experiments show that the proposed motion adaptation method guarantees the safety of the robots when the reference generated from human commands violated the physical limits of the robots.

The trade-off of the admittance and impedance behaviors is critical to the success and robustness of coordinated tasks. Admittance controllers will drift due to model errors, while impedance controllers cannot track the desired interaction force. The experimental result of moving a stack of books shown in Fig. \ref{fig:exp_direct_books} indicates that the adaptive trade-off between admittance and impedance behaviors increases the robustness of the coordinated motions in presence of model inaccuracy.

We conducted two long-horizon complex manipulation tasks to validate the capability of the proposed collaborative framework: the multi-operator part assembly and the teleoperated industrial connector insertion. 

Fig. \ref{fig:part_assembly} shows the multi-operator part assembly task completed by the collaboration of a remote operator, a local operator and a dual-arm robot. The remote operator first teleoperated the two robotic arms using the independent mode to approach the part on the conveyor belt and then switched to the bimanual mode to maneuver it over the other part placed on the table. Then the task was handed over to the local operator, who interactively guided and adjusted the fine manipulation process of the pose of the part by physically interacting with it using more detailed contact information. 

Fig. \ref{fig:connector_insertion} shows the connector insertion task, where two customized claw-like adaptors were designed for the manipulation of a cylindrical connector. The connector insertion task includes three sub-tasks: opening the lid of the socket which is partially covered by pebbles, unplugging the connector from an existed socket and inserting the connector to the new socket. Firstly one robotic arm was teleoperated to reach the socket covered by pebbles and open its lid. The dual-arm robot was then teleoperated to reach and grasp the connector and unplug it from the socket on the table. At last, the robot was teleoperated to maneuver the connector to approach the socket in the pebbles and completed the insertion task. 

The success of the part assembly task validated that the proposed framework has achieved guaranteed stability of reliable human-robot co-manipulation. The connector insertion task further validated that the proposed framework is compatible to dual-arm robots with additional customized adaptors for more dexterous grasping and manipulation.

\section{Conclusions}\label{sec:conclusions}
This paper presents a collaborative bimanual manipulation framework using optimal motion adaptation. The human Cartesian commands from teleoperation and physical interaction as well as the generated references by the admittance controller for the desired contact forces are optimized by Task-Space SEIKO to satisfy the physical limits of the robots and the contact constraints. The desired motions are realized by the interaction controller which combines independent controllers to generate the desired joint torque commands to control the two robotic arms. Experimental results validate that the proposed task-space formulation of SEIKO guarantees the safety during interactive and collaborative bimanual manipulation. The usage of the customized claw-like end-effectors of the Franka robotic arms has achieved the success of bimanual manipulation tasks for the insertion task of the cylindrical connector. In the future, we will explore the possibility of applying the motion adaption method with customized and/or commercial grippers for dexterous manipulation tasks towards manufacturing and medical applications.

\section*{Acknowledgment}

This research is supported by the  EPSRC Future AI and Robotics for Space (EP/R026092/1), ORCA (EP/R026173/1), NCNR (EP/R02572X/1), EU Horizon2020 project THING (ICT-2017-1), and EU Horizon2020 project Harmony (101017008).


\bibliographystyle{templates/IEEEtran}
\balance
\bibliography{ShortlistedReferences}

\end{document}